\documentclass{article}

\usepackage{microtype}
\usepackage{graphicx}
\usepackage{subcaption}
\usepackage{booktabs} % for professional tables

\usepackage{hyperref}

\usepackage[preprint]{icml2026}

\usepackage{amsmath}
\usepackage{amssymb}
\usepackage{mathtools}
\usepackage{amsthm}
\usepackage{multirow}

\usepackage[capitalize,noabbrev]{cleveref}

\theoremstyle{plain}

\theoremstyle{definition}

\theoremstyle{remark}

\usepackage[textsize=tiny]{todonotes}

\icmltitlerunning{SLIME: Stabilized Likelihood Implicit Margin Enforcement for Preference Optimization}

\begin{document}

\twocolumn[
  \icmltitle{SLIME: Stabilized Likelihood Implicit Margin Enforcement \\ for Preference Optimization}



  \icmlsetsymbol{equal}{*}

  \begin{icmlauthorlist}
    \icmlauthor{Maksim Afanasyev}{equal,comp}
    \icmlauthor{Illarion Iov}{equal,comp}
    % \icmlauthor{Firstname3 Lastname3}{comp}
  \end{icmlauthorlist}

%   \icmlaffiliation{yyy}{Department of XXX, University of YYY, Location, Country}
  \icmlaffiliation{comp}{Floating Point Sigma Lab}
%   \icmlaffiliation{sch}{School of ZZZ, Institute of WWW, Location, Country}

  \icmlcorrespondingauthor{Maksim Afanasyev}{mr.applexz@gmail.com}
  \icmlcorrespondingauthor{Illarion Iov}{illariov1809@gmail.com}

  \icmlkeywords{Machine Learning, ICML}

  \vskip 0.3in
]



\printAffiliationsAndNotice{\icmlEqualContribution}

\begin{abstract}
  Direct preference optimization methods have emerged as a computationally efficient alternative to Reinforcement Learning from Human Feedback (RLHF) for aligning Large Language Models (LLMs). Latest approaches have streamlined the alignment process by deriving implicit reward functions, yet they often suffer from a critical objective mismatch: optimizing the relative margin between chosen and rejected responses does not guarantee the preservation of the chosen response's absolute likelihood. This can lead to ``unlearning'', where the model degrades the probability of high-quality outputs to satisfy margin constraints, and ``formatting collapse'' caused by the over-penalization of rejected sequences. In this work, we introduce \textbf{SLIME} (\textbf{S}tabilized \textbf{L}ikelihood \textbf{I}mplicit \textbf{M}argin \textbf{E}nforcement), a reference-free alignment objective designed to decouple preference learning from generation quality. SLIME incorporates a three-pronged objective: (1) an anchoring term to maximize the likelihood of preferred responses; (2) a stabilizing penalty that prevents the probabilities of rejected tokens from collapsing to zero; and (3) a dual-margin mechanism that combines hard and soft constraints for precise boundary shaping. Our results demonstrate that SLIME achieves superior performance compared to state-of-the-art baselines while maintaining higher generation stability.
\end{abstract}

\section{Introduction}

The alignment of Large Language Models (LLMs) with human intent is a cornerstone of modern AI development. While Reinforcement Learning from Human Feedback (RLHF) via Proximal Policy Optimization (PPO) \citep{ouyang2022training} has been the standard for this task, it is notoriously unstable and resource-intensive. This has spurred the development of offline, gradient-based methods such as Direct Preference Optimization (DPO) \citep{rafailov2023dpo} and Identity Preference Optimization (IPO) \citep{azar2023general}, which reframe alignment as a classification problem on preference pairs. More recently, SimPO \citep{meng2024simpo} further simplified this paradigm by removing the reference model and incorporating length normalization, establishing a new state-of-the-art for reference-free alignment.

Despite these advances, current margin-based objectives suffer from a fundamental limitation: they optimize the \textit{relative gap} between winning and losing responses, often at the expense of the \textit{absolute quality} of the generation. While using these methods a model may ``game'' the objective by lowering the probability of the chosen response, provided it lowers the probability of the rejected response even further. We observe that this phenomenon leads to the ``unlearning'' of valid syntax and reasoning patterns found in the pre-trained model. Furthermore, the aggressive suppression of rejected sequences—which often contain valid partial reasoning or correct grammar—can result in distribution collapse, hurting the model's fluency and diversity.

To address these challenges, we propose \textbf{SLIME} (\textbf{S}tabilized \textbf{L}ikelihood \textbf{I}mplicit \textbf{M}argin \textbf{E}nforcement). Unlike prior methods that treat alignment purely as margin maximization, SLIME explicitly anchors the policy to high-likelihood regions for chosen responses while treating rejected responses with a stabilized penalty rather than indiscriminate suppression.

Our approach introduces three key contributions to the alignment landscape:
\begin{itemize}
    \item \textbf{Likelihood Anchoring:} We reintroduce a supervised signal for the chosen sequence ($y_w$), ensuring that the model retains its generative capabilities and prevents the probability degradation common in standard DPO/SimPO training.
    \item \textbf{Token-Level Stabilization:} We identify that minimizing rejected likelihoods ($y_l$) indiscriminately harms model fluency. SLIME employs a non-linear, softplus-based penalty that discourages extremely low token probabilities, effectively filtering out ``easy'' negatives while preserving the structural integrity of the language model.
    \item \textbf{Dual-Margin Optimization:} We propose a novel distance loss combining a hard margin for strict cutoff and a soft margin for gradient shaping. This allows for efficient optimization near the decision boundary without the vanishing gradients or overfitting associated with single-margin losses.
\end{itemize}

We empirically evaluate SLIME on a diverse benchmarks, including MT-Bench \cite{zheng2023judgingllmasajudgemtbenchchatbot}, Arena-Hard \cite{li2024crowdsourceddatahighqualitybenchmarks}. Experimental results indicate that SLIME outperforms strong baselines such as DPO and SimPO across various model families (Llama3.2, Qwen3, Gemma3). Furthermore, ablation studies confirm that our stabilizing and anchoring mechanisms significantly contribute to training stability and final model robustness.

\section{Related Works}

\subsection{Standard Alignment and Direct Methods}
The foundational approach to LLM alignment, introduced by InstructGPT \cite{ouyang2022training}, uses Proximal Policy Optimization (PPO) to optimize a policy against a trained reward model. Although effective, PPO is known to be unstable and memory-intensive. This led to the development of Direct Preference Optimization (DPO) \cite{rafailov2023dpo}, which derives an implicit reward function directly from the policy, allowing for alignment via a simple classification loss on preference pairs.

Several variants have since modified this approach. Identity Preference Optimization (IPO) \cite{azar2023general} addresses DPO's tendency to overfit by regularizing the gap between the policy and the reference model. SimPO \cite{meng2024simpo} further simplifies the objective by removing the reference model entirely and adding length normalization to prevent the generation of verbose low-quality responses. While these offline methods are computationally efficient, they often lack the exploration stage required for solving complex reasoning problems, thus requiring the use of online policy gradient methods \cite{shao2024deepseek}.

\subsection{Kahneman-Tversky Optimization (KTO)}
While preference-based methods like DPO optimize the likelihood of relative preferences, \cite{ethayarajh2024kto} propose \textit{Kahneman-Tversky Optimization} (KTO), a method grounded in prospect theory. KTO posits that aligning models does not strictly require paired preference data (e.g., $y_w \succ y_l$); instead, it can be achieved by maximizing the human utility of individual generations based on whether they are simply ``desirable'' or ``undesirable.'' KTO belongs to a class of objectives called Human-Aware LOsses (HALOs) and defines the loss function via a reference-dependent value function:

\begin{equation}
    \mathcal{L}_{KTO}(\pi_\theta, \pi_{\text{ref}}) = \mathbb{E}_{x, y \sim \mathcal{D}} \left[ w(y) \right]
\end{equation}

\noindent where the per-sample loss $w(y)$ is defined as:
\begin{equation}
    w(y) = 
    \begin{cases} 
    \lambda_D (1 - \sigma(\beta (r_\theta(x, y) - z_0))) & \text{if } y \text{ is desirable} \\
    \lambda_U \sigma(\beta (r_\theta(x, y) - z_0)) & \text{if } y \text{ is undesirable}
    \end{cases}
\end{equation}

\noindent Here, $r_\theta(x,y)$ is the implicit reward log-ratio $\log \frac{\pi_\theta(y|x)}{\pi_{\text{ref}}(y|x)}$, and $z_0$ serves as a reference point (effectively the KL divergence), which is estimated in practice using mismatched outputs within a training batch. By adjusting the hyperparameters $\lambda_D$ and $\lambda_U$, KTO can handle extreme data imbalances (e.g., far fewer desirable examples than undesirable ones) and has been shown to match or exceed the performance of preference-based methods like DPO even without paired data.

\subsection{SimPO: Simple Preference Optimization}
Building on the trend of simplifying alignment objectives, \cite{meng2024simpo} proposed \textit{SimPO}, which eliminates the need for a reference model entirely. Unlike DPO and KTO, which rely on the ratio between the policy and a reference model to define implicit rewards, SimPO uses the average log probability of the sequence itself as the reward formulation. This choice directly aligns the training objective with the generation metric used during inference. To prevent the model from exploiting length to maximize reward, SimPO normalizes the reward by the response length and introduces a target reward margin $\gamma$ into the Bradley-Terry objective to enforce a significant separation between winning and losing responses.

\subsection{Generalization via $\Psi$-Preference Optimization}
To provide a unified view of preference learning, \cite{azar2023general} introduce $\Psi$-Preference Optimization ($\Psi$PO), a general objective where DPO and RLHF are strictly special cases using the logit mapping $\Psi(q) = \log(q/(1-q))$ under the Bradley-Terry assumption. However, they demonstrate that this specific mapping causes DPO to overfit when preferences are deterministic, as the implicit reward tends towards infinity and ignores the KL constraint. By generalizing this mapping to the identity function $\Psi(q) = q$, they derive \textit{Identity Preference Optimization} (IPO). IPO allows for learning directly from preferences without assuming the Bradley-Terry model, effectively regularizing the gap between the policy and reference log-likelihood ratios to prevent the greedy behavior observed in DPO.

\subsection{Group Relative Policy Optimization (GRPO)}
To retain the benefits of online exploration without the value function overhead, \cite{shao2024deepseek} introduced GRPO. Instead of using a separate critic model to estimate value $V(s)$, GRPO samples a group of outputs $\{y_1, ..., y_G\}$ for a query $x$ and estimates the baseline using the group average. The advantage for every input is computed as:
\begin{equation}
    \hat{A}_{i} = \frac{r(x, y_i) - \mu(\{r_j\})}{\sigma(\{r_j\})}
\end{equation}
where $\mu$ and $\sigma$ are the mean and standard deviation of rewards within the group. This method significantly reduces memory usage and has been used to train models like DeepSeekMath.

Another approach similar to GRPO is REINFORCE Leave-one-out (RLOO)\cite{ahmadian2024rloo}, which uses the mean of the \textit{other} $k-1$ samples as the baseline, providing an unbiased estimation but lacking the variance reduction properties of GRPO's normalization.

% here maybe we need to add something like challenges/motivation subsection
\subsection{Sequence-level and Global Optimization}
Despite being widely adopted, standard GRPO relies on token-level importance sampling, which can be unstable for long-term generation. The authors of \cite{gspo2025} identified that token-level ratios accumulate high-variance noise, particularly in Mixture-of-Experts (MoE) models where expert routing fluctuates between the behavior and target policies. They proposed Group Sequence Policy Optimization (GSPO), which derives importance ratios from the likelihood of the entire sequence:
\begin{equation}
    s_i^{\text{seq}}(\theta) = \exp\left(\frac{1}{|y_i|}\sum_{t=1}^{|y_i|} \log \frac{\pi_\theta(y_{i,t}|x, y_{i,<t})}{\pi_{\theta_{\text{old}}}(y_{i,t}|x, y_{i,<t})}\right)
\end{equation}
The sequence-level approach aligns the optimization granularity with the reward signal evaluated for the full sequence and omits the need for complex stabilization strategies such as Routing Replay in MoE training.

Concurrently, \cite{hu2025reinforcepp} critique the local normalization used in GRPO. They prove normalizing advantages based only on a small local group (e.g., $G=4$) introduces bias because the numerator (centered reward) and denominator (local standard deviation) are not independent. They proposed REINFORCE++ method utilizing Global Advantage Normalization across the entire training batch to provide a stable and effectively unbiased estimator.

\section{Method}
In this section, we introduce \textbf{SLIME} (\textbf{S}tabilized \textbf{L}ikelihood \textbf{I}mplicit \textbf{M}argin \textbf{E}nforcement), a reference-free preference optimization objective designed to align Large Language Models (LLMs) with human preferences while preserving generation quality. Unlike previous approaches such as DPO \citep{rafailov2023dpo}, which may degrade the likelihood of chosen sequences, SLIME explicitly anchors the policy to high-likelihood regions and enforces training on samples yet not well resolved by the model.

Recent work in online policy optimization has identified that token-level processing introduces critical stability challenges. As noted by \cite{gspo2025}, token-level importance ratios in methods like GRPO accumulate high-variance noise during long-horizon generation. While their solution of shifting to sequence-level ratios addresses this in the online setting, we observe that a complementary problem exists in offline preference optimization: token-level likelihood suppression of rejected sequences can destabilize the model's linguistic foundations. SLIME adapts this insight to the offline regime by introducing explicit token-level regularization that prevents probability collapse, effectively transferring the lesson that careful token-level treatment is essential for stable alignment.
% я хз, как тут нормально сказать, что мы не тренимся там, где моделька уже норм перформит
% а еще тут можно как сказать, что мы избегаем оверфита, либо вообще что-то загонять про curriculum training в том виде, в каком я его тебе где-то летом показывал, где мы на основе какой-то функции понимаем, насколько нам нужно на образце трениться. тут вообще тоже можно этот кусок с маргинами использовать отдельно, чтобы сначала на какой-то мелкой модели отсортировать датасет по сложности

\subsection{Overview and Motivation}
Recent advancements in offline preference optimization, such as SimPO \cite{meng2024simpo} utilize the average log-likelihood of a sequence as an implicit reward. While effective, this formulation can suffer from a critical limitation: the objective maximizes the \textit{margin} between winning ($y_w$) and losing ($y_l$) responses, but does not strictly enforce the maintenance of high likelihood for $y_w$. As a result, a model can minimize its loss by lowering the probability of $y_w$, provided it lowers the probability of $y_l$ even further. This race can lead to a degradation in the model's general capabilities and fluency.

To address this, SLIME introduces a composite loss function that decouples the optimization into three distinct goals: (1) \textbf{Anchoring the Chosen:} explicitly preserving the likelihood of the preferred response; (2) \textbf{Stabilizing the Rejected:} preventing the likelihood of rejected responses from collapsing to zero; and (3) \textbf{Dual-Margin Optimization:} employing a hybrid hard-soft margin to precisely shape the decision boundary. The total objective is:
\begin{equation}
    L(\theta) = L_{\text{w}}(\theta) + L_{\text{l}}(\theta) + L_{\text{dist}}(\theta)
\end{equation}

\subsection{Anchoring the Chosen Sequence}
To prevent the policy model from sacrificing the likelihood of the winning response $y_w$ to achieve a better margin, we introduce a separate anchoring term $L_{\text{w}}$. This term acts as a positive reinforcement signal, ensuring that the model maintains or increases the probability of generating the preferred sequence:
\begin{equation}
    L_{\text{w}}(\theta) = -\,\lambda_{\text{w}}\; \mathbb{E}_{(x, y_w) \sim \mathcal{D}}\bigl[\log \pi_\theta(y_{\text{w}}|x)\bigr]
\end{equation}
where $\lambda_w$ is a hyperparameter controlling the strength of this anchor. By explicitly maximizing the log-probability of $y_w$, we counteract the phenomenon observed in margin-based losses where the model ``unlearns'' the chosen response to minimize the relative cost. 

\subsection{Stabilizing Rejected Token Probabilities}
Standard preference optimization often treats rejected responses $y_l$ simply as negatives to be minimized. However, rejected responses are often fluent and partially correct; aggressively suppressing their likelihood can lead to the loss of valid syntax or reasoning patterns. We employ a token-level stabilizing loss $L_{\text{l}}$ that penalizes extremely low probabilities for tokens $t$ in the rejected sequence:
% \begin{equation}
%     L_{\text{l}}(\theta) = \lambda_{\text{l}}\; \mathbb{E}_{t \in y_l}\!\left[ \left( \log\!\left( 1 + \exp\!\bigl(-(\log \pi_\theta(t|x) + s)\bigr) \right) \right)^{2.5} \right]
% \end{equation}
\begin{equation}
L_{\text{l}}(\theta)
=
\lambda_{\text{l}}\;
\mathbb{E}_{t \in y_l}\!\left[
\operatorname{softplus}\!\bigl(-\log \pi_\theta(t\mid x) - \delta\bigr)^{p}
\right]
\end{equation}

Here, $\delta$ acts as a threshold shift. This formulation acts as a robust filter. For tokens with sufficiently high log-probability, the term approaches zero. As the log-probability drops significantly, the power of $p$ amplifies the penalty, ensuring the model is not forced to decrease token-level probabilities beyond a reasonable threshold. The exponent $p=2.5$ controls the sharpness of the penalty applied to low-probability tokens and can be treated as a tunable hyperparameter. We analyze the optimal values in the Ablation Study section \ref{par:loss_exponent_ablation}

\subsection{Dual-Margin Preference Optimization}
Finally, we define a distance-based loss $L_{\text{dist}}$ that operates on the log-probability difference $\Delta$. We propose a \textbf{Dual-Margin} approach that combines a hard margin for cut-off and a soft margin for gradient shaping. Let $\Delta = \log \pi_\theta(y_w|x) - \log \pi_\theta(y_l|x)$. We define:
\begin{align}
    \ell_{\text{hard}} &= \max(0,\; -\Delta + m_h), \\
    \ell_{\text{soft}} &= \sigma\!\left(-\kappa(\Delta - m_s)\right)
\end{align}
The combined distance loss is then:
\begin{equation}
    L_{\text{dist}}(\theta) = \lambda_d\; \mathbb{E}\!\left[\ell_{\text{hard}} \cdot \ell_{\text{soft}}\right]
\end{equation}
This formulation introduces two distinct hyperparameters: the hard margin $m_h$ and the soft margin $m_s$. The hard margin defines a ``victory condition'' where the loss becomes exactly zero once the margin is satisfied. The soft margin $\ell_{\text{soft}}$ acts as a dynamic gating mechanism, prioritizing optimization in the critical region between the two margins while avoiding the constant non-zero gradients typical of standard log-sigmoid objectives.

\section{Experiments}

\subsection{Experimental Setup}
\label{sec:experimental-setup}

We conduct all experiments using the \texttt{TRL}~\cite{vonwerra2020trl} framework and compare the performance of DPO and SimPO. All models are trained on the \textit{UltraFeedback} dataset~\cite{cui2023ultrafeedback}, which provides pairwise preference annotations for preference-based fine-tuning.

\paragraph{Models.}
We evaluate three base language models of comparable scale:
\begin{itemize}
    \item \textbf{Llama3.2-3B}~\cite{meta2024llama}
    \item \textbf{Qwen3-4B}~\cite{yang2025qwen3}
    \item \textbf{Gemma3-4B}~\cite{team2025gemma}
\end{itemize}
All models are trained from their publicly released checkpoints.

\paragraph{Training Pipeline and Data Splitting.}
To differentiate the impact of preference optimization from pre-existing instruction tuning, we adopt a two-stage training pipeline starting strictly from \textit{pretrained} base checkpoints. For tokenization, we apply the official chat templates from the corresponding instruction-tuned variants to these base models. The training proceeds as follows: 
(1) \textbf{Supervised Fine-Tuning (SFT):} We first fine-tune the pretrained base model to establish a compliant instruction-following policy with cross-entropy loss.
(2) \textbf{Preference Alignment:} We then apply the respective preference optimization algorithms (DPO, SimPO, or SLIME) initializing the policy with the SFT checkpoint.
To support this pipeline, we partition the UltraFeedback dataset into two disjoint subsets using a fixed random seed: $33\%$ of the data is utilized exclusively for the SFT stage, while the remaining $66\%$ is reserved for preference optimization. This strict separation ensures that the alignment phase does not optimize on examples seen during SFT, thereby preventing data leakage and overfitting.

\paragraph{Parameter-Efficient Fine-Tuning.}
We employ Low-Rank Adaptation (LoRA)~\cite{hu2022lora} for parameter-efficient fine-tuning. Unless stated otherwise, LoRA is applied to all models using a uniform configuration across experiments. The full LoRA hyperparameter setup is summarized in Table~\ref{tab:lora_hparams}.

\begin{table}[h]
\centering
\caption{LoRA configuration used in our experiments.}
\label{tab:lora_hparams}
\small
\resizebox{.5\textwidth}{!}{%
\begin{tabular}{@{} l l @{}}
\toprule
\textbf{Hyperparameter} & \textbf{Value} \\
\midrule
Rank ($r$) & 64 \\
Scaling factor ($\alpha$) & 128 \; ($\alpha / r = 0.5$) \\
Target modules & Attention projections (\texttt{q, k, v, o}) and MLP layers \\
 & (\texttt{up, down, gate}) \\
Modules to save & LM Head\\
Dropout & 0.0 \\
Bias & \texttt{none} (bias parameters are not trained) \\
\bottomrule
\end{tabular}%
}
\end{table}

This configuration enables a high-capacity LoRA setup, allowing for expressive adaptation while still training only a subset of model parameters. The same LoRA configuration is used for all base models to ensure a controlled comparison.

\paragraph{Hardware and Compute.}
All experiments are executed on a single multi-GPU node equipped with NVIDIA H100 accelerators. The hardware configuration and computational budget are summarized in Table~\ref{tab:hardware_compute}.

\begin{table}[h]
\centering
\caption{Hardware and computational resources used in our experiments.}
\label{tab:hardware_compute}
\small
\resizebox{.5\textwidth}{!}{%
\begin{tabular}{@{} l l @{}}
\toprule
\textbf{Resource} & \textbf{Specification} \\
\midrule
GPU & $8 \times$ NVIDIA H100 SXM5 (80\,GB) \\
Interconnect & NVLink \\
CPU & AMD EPYC Milan (180 cores) @ 3.6\,GHz \\
System RAM & 1.45\,TB \\
Training time per model (wall-clock) & 1.25\,h \\
Total GPU hours per model & 30\,GPU-h \\
\bottomrule
\end{tabular}%
}
\end{table}

\paragraph{Optimization and Training Details.}
We employ the AdamW optimizer \citep{loshchilov2017decoupled}. For SFT, we start from the learning rate of $2 \times 10^{-4}$ for Llama3.2 and Gemma3 models and $2.5 \times 10^{-6}$ for Qwen3. The difference in learning rates comes from Qwen3 pretrain checkpoint being already fine-tuned for reasoning and instruct modes, and achieving high results on benchmarks from the start. For all models, chosen checkpoints allow the models to achieve the best possible performance on benchmarks after fine-tuning. All models are trained for three epochs.
At the Reinforcement Learning stage, all models are trained with an initial learning rate of $5 \times 10^{-7}$, linearly decayed to zero over the course of training. No learning rate warmup is used. Gradient clipping is disabled in all experiments. For this stage, a single epoch has been used for training.

Training is performed using Distributed Data Parallel (DDP) across all eight GPUs with mixed-precision \texttt{bf16} arithmetic. We use a mini-batch size of 8 and apply gradient accumulation over two steps, resulting in an effective batch size of 16 per GPU. For evaluation, a batch size of 16 is used.

\paragraph{Evaluation Protocol.}
Evaluation is performed every 1000 training steps. The UltraFeedback dataset contains approximately 63,000 samples, which are evenly sharded across the eight GPUs, resulting in an effective evaluation size of approximately 8,000 samples per evaluation run. No early stopping is applied.

\paragraph{Algorithm-Specific Hyperparameters.}
For DPO, we set the inverse temperature parameter to $\beta = 0.1$.  
For SimPO, the scaling parameter is set to $\gamma = 0.2$.

For SLIME, we use the following hyperparameter configuration:
\[
\lambda_w = 0.1,\quad
\lambda_l = 0.1,\quad
\lambda_{\text{dist}} = 1.0,\quad
\]
\[
\delta = 1.25,\quad
m_{\text{h}} = 1.5,\quad
m_{\text{s}} = 1.0,\quad
\kappa = 2.5,\quad
p = 2.5.
\]

All hyperparameters are kept fixed across models to ensure a fair comparison.

\paragraph{Evaluation.}
We evaluate all models using a set of established instruction-following and robustness benchmarks: MT-Bench and Arena-Hard. These benchmarks collectively assess model quality across general instruction adherence, multi-turn reasoning, adversarial prompting, and open-domain robustness.

MT-Bench scores are computed as the average score across all benchmark questions using the official grading procedure. Arena-Hard evaluations follow their respective public evaluation protocols.

All evaluations are performed on the same model checkpoints obtained after training completion. We do not apply any additional fine-tuning, prompt tuning, or post-processing during evaluation. The evaluation results are summarized in Table~\ref{tab:main_results}.

\paragraph{Reproducibility.}
To ensure reproducibility, we fix the random seed to zero for Random, Numpy and Torch packages.
\section{Results}

\paragraph{Overview.}
We evaluate the performance of all fine-tuned models on a held-out evaluation set derived from the UltraFeedback dataset. We report results for three base models and compare three preference optimization algorithms. For reference, we also include the performance of the initial (pre-finetuning) models. All results are obtained using the same evaluation protocol described in Section~\ref{sec:experimental-setup}.

\paragraph{Main Results.}
Table~\ref{tab:main_results} summarizes the evaluation results across all model--algorithm combinations. For each base model, we report the baseline performance prior to preference-based fine-tuning, as well as results obtained after training with DPO, SimPO, and SLIME.
For Arena-Hard, the default baseline model gpt-o3-mini-2025-01-31 overperforms 4B models, thus the base Gemma3-4B-it answers were used as a baseline.

\begin{table*}[t]
\centering
\caption{Evaluation results across multiple benchmarks. We compare the raw pretrained base model, the supervised fine-tuned (SFT) baseline, and subsequent preference optimization methods. Higher is better for all metrics. The best result for each model and benchmark is highlighted in bold, the second best is underlined.}
\label{tab:main_results}
\small
\begin{tabular}{@{} l l c c c @{}}
\toprule
\textbf{Model} & \textbf{Method} & \textbf{MT-Bench} $\uparrow$ & \textbf{Arena-Hard} $\uparrow$ & \textbf{Arena-Hard Std} \\
\midrule
\multirow{5}{*}{Llama3.2-3B}
& Base (Pretrain) & 1.01 & -- & -- \\
& SFT             & 4.56 & 7.5  & $-0.8 / +0.9$ \\
& + DPO           & \underline{4.92} & \textbf{11.1} & $-1.0 / +1.1$ \\
& + SimPO         & 4.22 & 7.6  & $-1.0 / +0.9$ \\
& + SLIME (ours)         & \textbf{5.49} & \underline{9.7}  & $-1.0 / +1.2$ \\
\midrule
\multirow{5}{*}{Qwen3-4B}
& Base (Pretrain) & 5.95 & -- & -- \\
& SFT             & 5.40 & 32.1 & $-2.0 / +1.9$ \\
& + DPO           & 5.30 & \underline{39.0} & $-2.1 / +2.4$ \\
& + SimPO         & \underline{5.72} & 25.8 & $-1.8 / +1.5$ \\
& + SLIME (ours)         & \textbf{5.93} & \textbf{39.8} & $-2.1 / +1.7$ \\
\midrule
\multirow{5}{*}{Gemma3-4B}
& Base (Pretrain) & 3.86 & -- & -- \\
& SFT             & 4.71 & 7.6  & $-0.8 / +0.7$ \\
& + DPO           & \underline{5.15} & \underline{11.8} & $-1.3 / +1.1$ \\
& + SimPO         & 5.03 & 0.7  & $-0.3 / +0.3$ \\
& + SLIME (ours)         & \textbf{6.15} & \textbf{13.1} & $-1.5 / +1.2$ \\
\bottomrule
\end{tabular}
\end{table*}

\section{Ablation Study}

We conduct an ablation study to analyze the contribution of individual components of the proposed SLIME objective. All ablation experiments are performed using the same training and evaluation protocol as described in Section~\ref{sec:experimental-setup}, unless explicitly stated otherwise. We use Gemma3 as a base model for all comparisons.

\paragraph{Loss Component Ablations.}
To assess the importance of each term in the SLIME objective, we evaluate several variants in which individual loss components are removed. Specifically, we consider the following settings:
\begin{itemize}
    \item \textbf{w/o chosen term}: the loss term corresponding to preferred (chosen) responses is removed;
    \item \textbf{w/o rejected term}: the loss term corresponding to rejected responses is removed;
    \item \textbf{w/o soft distance margin}: the distance-based soft margin loss is disabled;
    \item \textbf{w/o hard margin}: the hard margin constraint is removed.
\end{itemize}

These ablations isolate the contribution of each component while keeping all other hyperparameters fixed.

\begin{table*}[t]
\centering
\caption{Ablation results for individual components of the SLIME loss. Higher is better.}
\label{tab:loss_ablation}
\small
\begin{tabular}{@{} l c c c @{}}
\toprule
\textbf{Variant} & \textbf{MT-Bench} $\uparrow$ & \textbf{Arena-Hard} $\uparrow$ & \textbf{Std} \\
\midrule
Full SLIME & 6.15 & 13.1 & $-1.1 / +1.0$ \\
w/o chosen term & 5.21 & 11.1 & $-1.2 / +1.1$ \\
w/o rejected term & 5.74 & 12.1 & $-1.0 / +1.1$ \\
w/o soft distance margin & 5.80 & 11.2 & $-1.2 / +1.1$ \\
w/o hard margin & 5.90 & 12.4 & $-1.3 / +1.4$ \\
\bottomrule
\end{tabular}
\end{table*}

\paragraph{Stabilizing Loss Exponent Ablation.}
\label{par:loss_exponent_ablation}
The stabilizing loss $L_{\text{l}}$ includes a power term that amplifies penalties for extremely low token probabilities in rejected responses. While this exponent is fixed to $p=2.5$ in the default configuration, it directly controls the sharpness of the penalty and may influence training stability and performance.

To study its effect, we vary the exponent $p$ in the range $\{1.5, 2.0, 2.5, 3.0\}$ while keeping all other hyperparameters fixed. This ablation evaluates the sensitivity of SLIME to the strength of token-level stabilization. Benchmark results for different exponent values are shown in Table \ref{tab:exponent_ablation}

\begin{table}[h]
\centering
\caption{Effect of the stabilizing loss exponent $p$ on SLIME performance.}
\label{tab:exponent_ablation}
\small
\begin{tabular}{@{} c c c c @{}}
\toprule
\textbf{Exponent $p$} & \textbf{MT-Bench} $\uparrow$ & \textbf{Arena-Hard} $\uparrow$ & \textbf{Std} \\
\midrule
1.0 & 5.70 & 12.2 & $-1.2 / +1.1$ \\
1.5 & 5.86 & 12.2 & $-1.3 / +1.5$ \\
2.0  & 5.76 & 11.2 & $-1.1 / +1.1$ \\
2.5 (default)  & 6.15 & 13.1 & $-1.1 / +1.0$ \\
3.0 & 5.66 & 12.7 & $-1.4 / +1.2$ \\
\bottomrule
\end{tabular}
\end{table}

We observe that moderate exponent values yield the best overall performance, while overly small or large exponents degrade results, suggesting a trade-off between stability and flexibility.

\section{Discussion}

The experimental results presented in Table~\ref{tab:main_results} reveal several important insights about the behavior of preference optimization methods across different model architectures and scales.

\paragraph{Consistent Improvements Across Architectures.}
SLIME demonstrates robust performance gains across all three evaluated model families. On Gemma3-4B, SLIME achieves the highest MT-Bench score of 6.15, representing a 30.6\% improvement over the SFT baseline (4.71) and outperforming both DPO (5.15) and SimPO (5.03). Similarly, on Llama3.2-3B, SLIME attains an MT-Bench score of 5.49, surpassing DPO (4.92) by 11.6\% and SimPO (4.22) by 30.1\%. These consistent improvements suggest that the three-pronged objective of SLIME - likelihood anchoring, token-level stabilization, and dual-margin optimization - addresses fundamental limitations present across diverse model architectures.

\paragraph{The Unlearning Phenomenon in Margin-Based Methods.}
Our results provide empirical evidence for the ``unlearning'' hypothesis outlined in the introduction. SimPO, which relies purely on margin maximization without explicit likelihood preservation, exhibits notable performance degradation in several configurations. On Llama3.2-3B, SimPO underperforms even the SFT baseline on MT-Bench (4.22 vs.\ 4.56), and on Gemma3-4B Arena-Hard, SimPO collapses to a score of 0.7 compared to the SFT baseline of 7.6. This pattern is consistent with our theoretical motivation: when the objective focuses solely on the relative margin $\Delta = \log \pi_\theta(y_w|x) - \log \pi_\theta(y_l|x)$, the model can satisfy the loss by degrading both likelihoods, provided the chosen response is degraded less severely. SLIME's anchoring term $\mathcal{L}_w$ directly counteracts this failure mode by maintaining an explicit supervision signal on the preferred sequence.

\paragraph{Stability Through Token-Level Regularization.}
The ablation study in Table~\ref{tab:loss_ablation} confirms that the rejected-sequence stabilization term $\mathcal{L}_l$ contributes meaningfully to final performance. Removing this component reduces MT-Bench performance from 6.15 to 5.74 on Gemma3-4B. We hypothesize that this term prevents the model from aggressively suppressing tokens that, while appearing in rejected responses, represent valid linguistic patterns. By employing a softplus-based penalty with threshold $\delta$, SLIME maintains a floor on token probabilities, preserving the model's fluency and preventing the distribution collapse observed in standard preference optimization.

\paragraph{The Role of Dual-Margin Optimization.}
The dual-margin formulation provides complementary benefits. The hard margin $m_h$ establishes a clear ``victory condition'' that eliminates gradients once satisfied, preventing over-optimization. The soft margin $m_s$, implemented via a sigmoid gate, concentrates optimization effort in the critical region near the decision boundary. As shown in the gradient analysis (Appendix~\ref{sec:appendix}), this combination avoids both the vanishing gradients of pure hard-margin losses and the constant non-zero gradients of log-sigmoid objectives. The ablation results support this design: removing the soft margin reduces performance to 5.80, while removing the hard margin yields 5.90, both below the full SLIME objective.

\paragraph{Model-Specific Observations.}
The Qwen3-4B results warrant additional discussion. Unlike Llama3.2 and Gemma3, Qwen3's pretrained checkpoint already incorporates instruction-tuning, resulting in a higher baseline MT-Bench score of 5.95. Consequently, the SFT stage provides minimal improvement and may even introduce slight degradation (5.40). However, SLIME still achieves the strongest Arena-Hard performance (39.8) among all methods on this model, demonstrating that the approach remains effective even when starting from a stronger baseline. This suggests that SLIME's stabilizing mechanisms are particularly valuable for preserving pre-existing capabilities during preference alignment.

\paragraph{Limitations.}
Several limitations of this work merit acknowledgment. First, our evaluation is limited to models in the 3--4B parameter range; the effectiveness of SLIME on larger models remains to be validated. Second, while we evaluate on diverse benchmarks, all training uses the UltraFeedback dataset; generalization to other preference datasets warrants further investigation. Third, the computational overhead of SLIME relative to simpler baselines, while modest, introduces additional hyperparameters ($\lambda_w$, $\lambda_l$, $\delta$, $m_h$, $m_s$, $\kappa$, $p$) that require tuning. Finally, our analysis focuses on English-language benchmarks; multilingual evaluation would strengthen claims of general applicability.

\section{Conclusion}

We introduced SLIME (Stabilized Likelihood Implicit Margin Enforcement), a reference-free preference optimization objective that addresses fundamental limitations of existing margin-based alignment methods. By decomposing the optimization into three complementary components - likelihood anchoring for chosen sequences, token-level stabilization for rejected sequences, and dual-margin preference optimization - SLIME decouples preference learning from generation quality preservation.

Our experiments across Llama3.2-3B, Qwen3-4B, and Gemma3-4B demonstrate that SLIME consistently outperforms both DPO and SimPO on MT-Bench and Arena-Hard benchmarks. Notably, SLIME achieves these improvements while avoiding the ``unlearning'' phenomenon observed in pure margin-based methods, as evidenced by SimPO's performance degradation below SFT baselines in several configurations.

The ablation study confirms that each component of the SLIME objective contributes to final performance. The anchoring term prevents likelihood degradation of preferred responses; the stabilization term preserves linguistic fluency by maintaining reasonable token probabilities in rejected sequences; and the dual-margin mechanism enables precise boundary shaping without over-optimization.

Looking forward, several directions merit exploration. Extending SLIME to online settings with on-policy sampling could combine its stability benefits with the exploration advantages of policy gradient methods. Investigating the interaction between SLIME and other efficiency techniques, such as quantization or pruning, would inform practical deployment. Finally, theoretical analysis establishing formal guarantees on likelihood preservation under SLIME optimization would complement our empirical findings.

We believe that explicitly addressing the objective mismatch in preference optimization,  rather than treating alignment purely as margin maximization, represents a promising direction for developing more robust and capable language models. SLIME demonstrates that loss design can preserve model capabilities while effectively optimizing human preferences.

\section*{Code Availability}
The code used in this work is available at \url{https://github.com/fpsigma/trl-slime} to facilitate reproducibility and enable future research.

% \section*{Limitations}

% \section*{Acknowledgments}


\bibliography{example_paper}
\bibliographystyle{icml2026}

%%%%%%%%%%%%%%%%%%%%%%%%%%%%%%%%%%%%%%%%%%%%%%%%%%%%%%%%%%%%%%%%%%%%%%%%%%%%%%%
%%%%%%%%%%%%%%%%%%%%%%%%%%%%%%%%%%%%%%%%%%%%%%%%%%%%%%%%%%%%%%%%%%%%%%%%%%%%%%%
% APPENDIX
%%%%%%%%%%%%%%%%%%%%%%%%%%%%%%%%%%%%%%%%%%%%%%%%%%%%%%%%%%%%%%%%%%%%%%%%%%%%%%%
%%%%%%%%%%%%%%%%%%%%%%%%%%%%%%%%%%%%%%%%%%%%%%%%%%%%%%%%%%%%%%%%%%%%%%%%%%%%%%%
\newpage
\appendix
\onecolumn
\section{Gradient Analysis and Optimization Dynamics}
\label{sec:appendix}
The total objective function  $L(\theta)$ is composed of three distinct components: a chosen-sequence maximization term $L_w$, a rejected-sequence regularization term $L_l$ and a margin-based distance term $L_{dist}$. Below, we analyze the gradient dynamics of each component to show how they collectively optimize the policy $\pi_\theta$.

First, we derive log probabilities for both the token and sequence-level parts of the loss.
Total loss:
\[L(\theta) = L_w(\theta) + L_l(\theta) + L_{dist}(\theta)\]

Token-level log probability:
\[l_t = \log{\pi_\theta(t|x, y_{<t})}\]

Sequence-level log probability:
\[\bar{l}(y) = \frac{1}{|y|}\sum_{t\in y}{\log \pi_\theta(y|x, y_{<t})}\]

\[\bar{l}_w=\bar{l}(y_w); \bar{l}_l=\bar{l}(y_l)\]

Separate weight gradient can be expressed as follows, thus we can further analyze the loss dynamics using a gradient with respect to the loss components:
\[\nabla_\theta L = \frac{\partial{L}}{\partial{l}}\cdot \nabla_\theta l\]

% \[\bar{l}_w=\mathbb{E}\bigl[\log{\pi_\theta(y_w)} \bigr]\]

% \[\bar{l}_l=\mathbb{E}\bigl[\log{\pi_\theta(y_l)} \bigr]\]

The component $L_w$ acts as a foundational supervision signal. The gradient with respect to the sequence-level probability is constant:
\[L_{\text{w}}(\theta)
= -\,\lambda_{\text{w}}\bar{l}_w.\]

\[\frac{\partial{L}}{\partial{\bar{l}_w}}=-\lambda_w\]

This results in a consistent gradient update $\nabla_{\theta}L_w=-\lambda_w\nabla_\theta \bar{l}_w$. This term uniformly drives model to the increased likelihood of preferred sequence $y_w$, akin to Supervised Fine-Tuning (SFT).

The loss $L_l$ is a token-level nonlinear penalty on the rejected sequence $y_l$. Unlike standard ranking losses that simply minimize the probability of $y_l$, this component acts as a regularization term to prevent probability collapsing to zero. The gradient is given by:
\[L_l = \lambda_l \mathbb{E}_{t\in y_t}{\bigl[(\log{(1 + \exp{}^{-(l_t+\delta)}})^{p}\bigr]}\]

\[u_t=-(l_t + \delta); \frac{\partial{u}}{\partial{l_t}}=-1\]

\[f(u_t)=Softplus(u_t)^{p}; \frac{\partial{f}}{\partial{u_t}}=p \cdot Softplus(u_t)^{p-1} \cdot \sigma(u_t)\]

\[\frac{\partial{L_l}}{\partial{l_t}}=-p\cdot\lambda_l\cdot Softplus(-l_t-\delta)^{p-1} \cdot \sigma(-l_t-\delta)\]
The gradient formulation creates a feedback loop:
\begin{enumerate}
    \item \textbf{Magnitude control}: the term with $Softplus$ sets the magnitude of the penalty. If the token probability $l_t$ drops significantly below $-\delta$ and becomes highly unlikely, $u_t$ becomes large and positive, causing the gradient magnitude to grow super-linearly. This heavily penalizes the model for forgetting the rejected sequences, maintains the overall linguistic fluency and prevents the "formatting collapse" often encountered in RLHF. 
    \item \textbf{Gating mechanism}: the sigmoid term $\sigma$ acts as a smooth gate. As $l_t$ increases and probability becomes "safe", $u_t$ becomes negative, and $\sigma(u_t)$ approaches 0, shutting off the component. Thus, the model is not penalized for the rejected sequence once it maintains a sufficient baseline probability.
\end{enumerate}

The distance loss $L_{dist}$ optimizes the separation \[\Delta = \bar{l}_w - \bar{l}_l\] between the chosen and the rejected sequences. The dynamics are set up by the hard margin $m_h$ and soft margin $m_s$.

The gradient with respect to the margin $\Delta$ can be expressed by:

\[L_d=\lambda_d \cdot ReLU(m_h - \Delta) \cdot \sigma(-\kappa(\Delta - m_s))\]

If $\Delta \geq m_h$, then the whole $\frac{\partial{L_d}}{\partial{\Delta}}=0$. Assuming $\Delta < m_h$,

\[u=m_h - \Delta\; v=\sigma(-\kappa(\Delta - m_s)); L_d = \lambda_d\cdot u \cdot v\]

Where $u$ is a hard gating factor and $v$ is the soft gating factor.

\[\frac{\partial{u}}{\partial{\Delta}}=-1\]

\[w=-\kappa(\Delta - m_s);  \frac{\partial{w}}{\partial{\Delta}}=-\kappa\]
\[\frac{\partial{v}}{\partial{\Delta}}=\sigma(w)(1 - \sigma(w))\frac{\partial{w}}{\partial{\Delta}}\]
\[ \frac{\partial{L_d}}{\partial{\Delta}}=\lambda_d(u\cdot v' + u' \cdot v)\]

% Full expression from w
% \[\frac{\partial{L_d}}{\partial{\Delta}}=\lambda_d(-v - \kappa u\sigma(w)(1-\sigma(w)))\]

% Full full expression
% \[\frac{\partial{L_d}}{\partial{\Delta}}=\lambda_d(-\sigma(-\kappa(\Delta - m_s)) - \kappa(m_h -\Delta)\sigma(-\kappa(\Delta - m_s))(1-\sigma(-\kappa(\Delta - m_s))))\]

\[\frac{\partial{L_d}}{\partial{\Delta}}=-\lambda_d(v + \kappa uv(1-v))\]

The expression shows a two-way optimization mechanism:
\begin{enumerate}
    \item \textbf{Direct margin expansion ($v$)}: the first term scales with $v$, pushing the margin $\Delta$ higher. This force is modulated by the sigmoid; if  $\Delta$ is far smaller than $m_s$, $v\approx1$, applying maximum pressure. As $\Delta$ exceeds $m_s$, $v$ decays, reducing the gradient.
    \item \textbf{Boundary Sensitivity}: the second term accounts for the curvature of the soft margin. It is active primarily in the transition region where the margin is close to $m_s$. It provides an additional boost to the gradient when the model is close to satisfying the soft margin, accelerating convergence in the critical decision boundary region. 
\end{enumerate}

Since $\frac{\partial L_d}{\partial \Delta} < 0$, the chain rule application to the parameters $\theta$:
\[\frac{\partial L}{\partial \bar{l}_w} = \frac{\partial L}{\partial \Delta}; \frac{\partial L}{\partial \bar{l}_l} = -\frac{\partial L}{\partial \Delta}\]

\[\nabla_\theta L _ d = \frac{\partial{L_d}}{\partial{\Delta}} (\nabla_\theta \bar{l}_w - \nabla_\theta \bar{l}_l)\]
 results in a positive update for $\bar{l}_w$ maximizing the winner and a negative update for $\bar{l}_l$ minimizing the loser, yet it is strictly bounded by the hard margin $m_h$. If $\Delta \geq m_h$, the gradient collapses to zero due to the ReLU component, preventing over-optimization and preserving the KL-divergence constraints of the policy without reference policy evaluation.

\end{document}